%% file: layout_estimation_iccv.tex
\documentclass[10pt,twocolumn,letterpaper]{article}
 
\usepackage{wacv}
\usepackage{times}
\usepackage{epsfig}
\usepackage{graphicx}
\usepackage{amsmath}
\usepackage{amssymb}
\usepackage{xcolor}
\usepackage{booktabs}
 \usepackage{paralist}

\def\wacvPaperID{****} 

\wacvfinalcopy 

\ifwacvfinal
\def\assignedStartPage{9876} 
\fi

\ifwacvfinal
\usepackage[breaklinks=true,bookmarks=false]{hyperref}
\else
\usepackage[pagebackref=true,breaklinks=true,colorlinks,bookmarks=false]{hyperref}
\fi

\ifwacvfinal
\else
\fi

\DeclareMathOperator*{\argmax}{arg\,max}

 \usepackage{siunitx} 
 
\begin{document}
 
\title{RoomStructNet: Learning to Rank Non-Cuboidal Room Layouts From Single View}

\author{%
	Xi Zhang, Chun-Kai Wang, Kenan Deng, Tomas Yago-Vicente, Himanshu Arora\\
	Visual Search \& AR\\
	Palo Alto, CA 94301 \\
	\tt\small{\{xizhn,ckwang,kenanden,victomas,arorah\}@amazon.com} 
}

\maketitle
 
\begin{abstract}
	In this paper, we present a new approach to estimate the layout of a room from its single image. While recent approaches for this task use robust features learnt from data, they resort to optimization for detecting the final layout. In addition to using learnt robust features, our approach learns an additional ranking function to estimate the final layout instead of using optimization. To learn this ranking function, we propose a framework to train a CNN using max-margin structure cost. Also, while most approaches aim at detecting cuboidal layouts, our approach detects non-cuboidal layouts for which we explicitly estimates layout complexity parameters. We use these parameters to propose layout candidates in a novel way. Our approach shows state-of-the-art results on standard datasets with mostly cuboidal layouts and also performs well on a dataset containing rooms with non-cuboidal layouts.

\end{abstract}
 
\input{sec_intro.tex}
\input{sec_representation.tex}

\input{sec_features}
\input{sec_proposals}

\input{sec_ranking}

\input{sec_results}
\input{sec_conclusion}

{\small
\bibliographystyle{ieee_fullname}
\bibliography{egbib}
}
 
\end{document}

%% file: sec_intro.tex
\section{Introduction} \label{sec:intro}

The problem of room layout estimation from a single image corresponds to identifying the extent of walls, floor, and ceiling of the depicted room as if it was empty. 
Room layout estimation is a fundamental component of indoor scene understanding that remains unsolved. Furthermore, it has recently gained great interest due to its direct application to holistic scene understanding~\cite{huang2018cooperative, huang2018holistic}, robotic navigation~\cite{IM2CAD} and augmented reality~\cite{gal2014flare, sra2016procedurally}. 

\begin{figure}[t]
	\centerline{
		\begin{tabular}{c}
			\resizebox{0.4\textwidth}{!}{\rotatebox{0}{
					\includegraphics{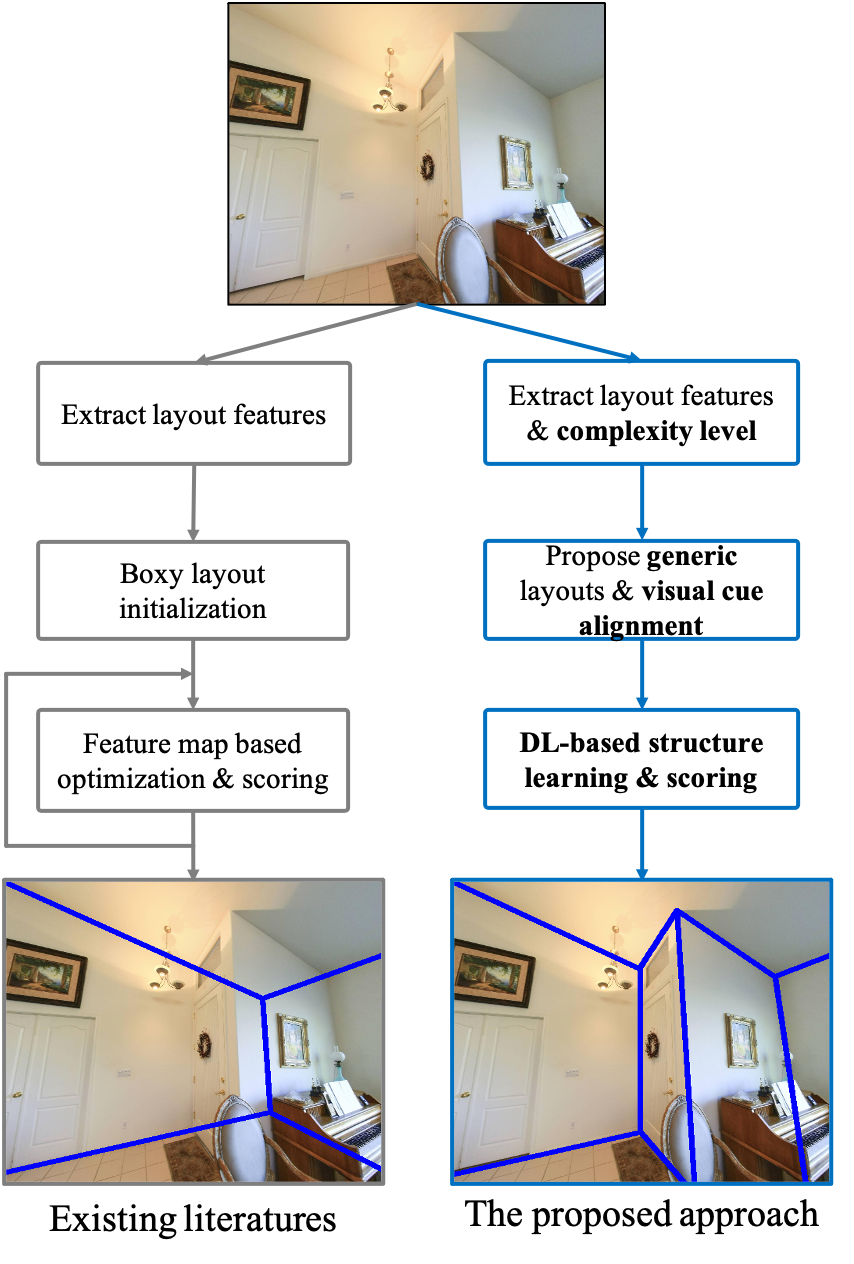}}}
			\\
	\end{tabular}}
	\caption{Comparison of layout prediction pipelines between existing literatures and the proposed approach. The proposed approach differs from existing literatures and has improvement at every step.  We use bold font to highlight key contribution and difference of the proposed approach at each step.}
	\label{fig: pipeline comparison}
\end{figure}

The first stage of most existing approaches is to detect features to localize the layout, see Figure~\ref{fig: pipeline comparison} left.
For instance, classical approaches use vanishing point aligned line segments~\cite{Hedau, Gupta}, superpixels~\cite{Zhang}, or edges~\cite{Mallya}. More recent works have replaced these bottom up image features with CNN based features learned from data, such as: informative edge maps~\cite{Mallya, Ren}, location of boundaries where two room surfaces meet (e.g. wall-wall boundaries)~\cite{PIONet,LayoutNet}, location of corners where three room surfaces meet (e.g. wall-wall-floor corner)~\cite{RoomNet,LayoutNet}, segmentation masks of room surfaces~\cite{IM2CAD,DasGupta,Ren} or depth maps of dominant planes~\cite{GeoLayout, RenderAndCompare}. In a similar manner, we propose to use room boundaries, corners, and surface labels as features to learn. We believe these have complementary information that provides robustness to clutter, which will allow us to estimate better layouts.

After feature detection, most approaches use the detected features to estimate a cuboidal layout, see Figure~\ref{fig: pipeline comparison} left. Assuming the room to have a box shape allows for easier parameterization and solution search. Many approaches explicitly estimate orthogonal vanishing points and enforce walls to be oriented along these~\cite{Hedau,Gupta,Mallya,Ren}. Others assume existence of at most 3 walls while estimating surface labels~\cite{DasGupta,Ren}. More recent approaches limit themselves only to layout configurations of visible walls, floor, and ceiling that can occur for a cuboidal layout~\cite{PIONet}. Some approaches even 
detect the configuration explicitly from a predefined small set of cuboidal room layout types~\cite{RoomNet}. It is worth noting that not all rooms can actually be modeled as a box. This is especially true for modern homes with multi-functional rooms like living room/kitchen area or master bedroom/bathroom area. Hence, we propose a generic room layout model that can also handle such rooms with arbitrary number of walls. Additionally, as part of our feature extraction we propose to train a network to find the parameters of our generic layout model that best suit the given input image, see Figure~\ref{fig: pipeline comparison} right. Moreover, we propose a novel sampling algorithm that generates plausible room layouts based on the estimated generic layout parameters and the attention provided by our robust learned features.

Recent approaches that attempted generalization to non-cuboidal layouts operate on panoramic images, whereas we operate on perspective images~\cite{LayoutNet, fernandez2020corners}. The work in~\cite{flat2Layout} uses a representation to detect non-cuboidal layouts which estimates the probability of wall-wall boundary locations and corners in a reprojected the image and uses it to sample layouts. Similarly,~\cite{RenderAndCompare, howard2018thinking} uses planes estimated by analyzing RGB or RGBD images to sample non-cuboidal layouts. In contrast to these works, our model explicitly outputs generic layout parameters such as number of visible walls and corners, instead of just sampling non-cuboidal layouts. 
   

\begin{figure*}[ht]
	\centerline{
		\begin{tabular}{c}
			\resizebox{0.9\textwidth}{!}{\rotatebox{0}{
					\includegraphics{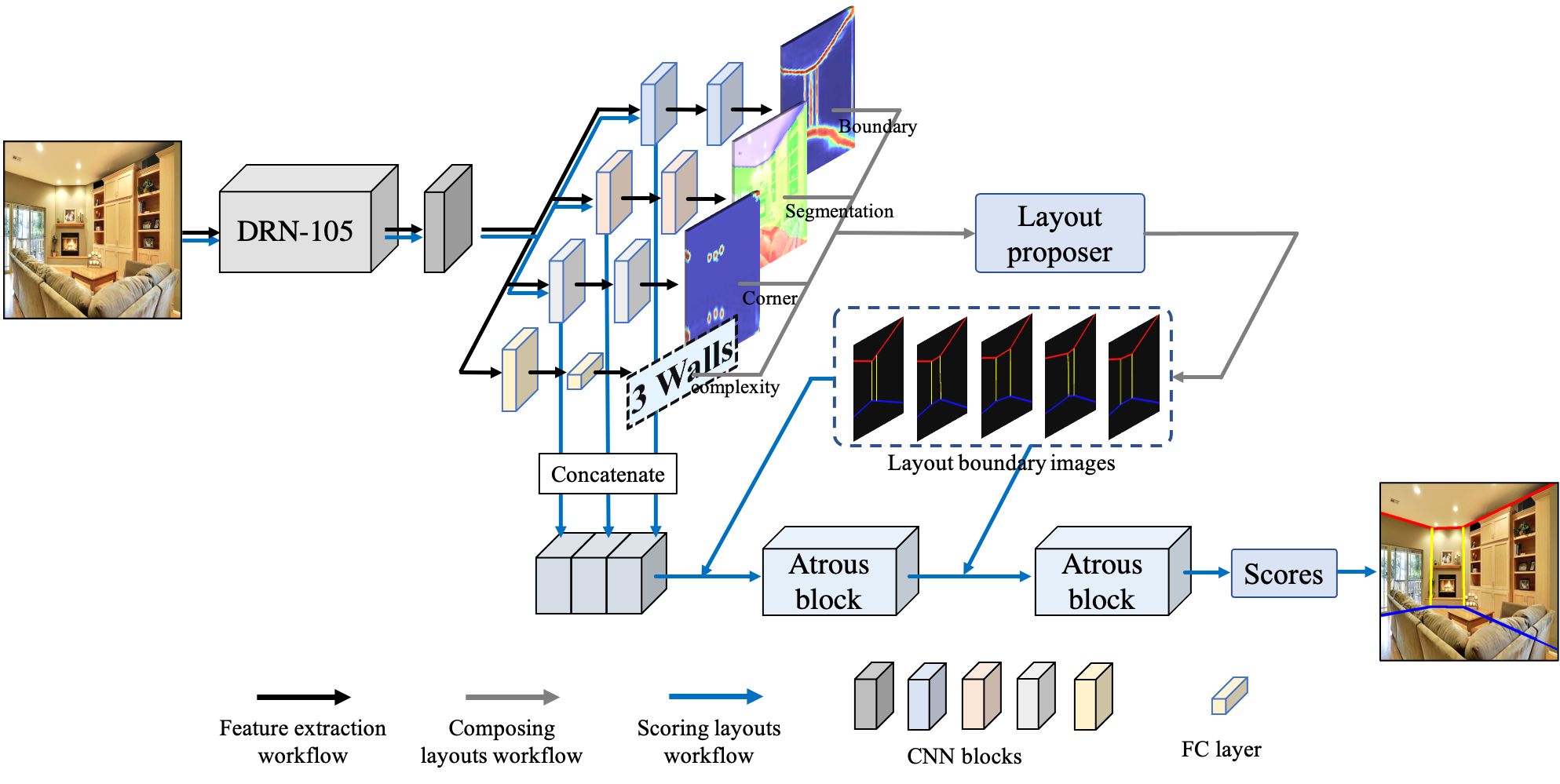}}}
			\\
	\end{tabular}}
	\caption{Our RoomStructNet pipeline. First, we predict different feature maps and the layout complexity level using a trained featured extraction network, whose process is shown by the black arrows. Second, we propose candidate layouts using the extracted features and the complexity level, shown by the grey arrows. Lastly, we score the candidate layout, shown by the blue arrow.}
	\label{fig: whole pipeline}
\end{figure*} 

In order to estimate the room layout, most approaches either use heuristics or an optimization procedure guided by the layout features extracted in previous stages, see Figure~\ref{fig: pipeline comparison} left. These can include aggregating detected features in layout regions~\cite{Ren}, using physics inspired gradient descent on layout boundary confidence maps~\cite{PIONet}, greedy descent of layout boundaries using layout surface label confidence maps~\cite{DasGupta} or optimization using combined 2D and 3D cost functions~\cite{LayoutNet,RenderAndCompare}. 
Layout structure is complex, it needs to satisfy many constraints such as perpendicularity of floor and walls, planar nature of each surface. Hence, it may require non-trivial combinations of multiple kinds of features. We do believe these combinations are best learned from data. For instance, to exploit the structured nature of a layout, the seminal work of~\cite{Hedau} uses max-margin structured output regression to train a layout ranker as a linear combination of different kinds of features. Instead of a linear combination, we propose a neural network to score layouts and rank them. In addition, we propose a novel way to train this neural network using max-margin structure cost. Comparing to methods~\cite{pintore2020atlantanet} estimating generic layouts from 360\si{\degree} image, the proposed approach address generic layout estimation by using a single perspective image.

To evaluate our proposed method, we conduct experiments on the challenging Hedau~\cite{Hedau} and LSUN~\cite{yu2015lsun} room layout benchmark dasasets. Our method outperforms the current state of the art approach of Hsian $\etal$~\cite{flat2Layout} in both datasets by a clear margin.  Note that these datasets contain only cuboidal layouts ground truth annotations. Therefore, we created a new dataset with rooms with generic non-cuboidal layouts to fully evaluate the potential of our method. This dataset consists of a large collection images (50K) extracted from panoramas of the PanoContext~\cite{zhang2014panocontext} dataset and relabeled by us. Our method obtains impressive qualitative results on this more challenging unconstrained data, while achieving decent quantitative performance. 

Our main contributions are:
\begin{compactenum}
	\item We propose a novel generic room layout estimation system that achieves the state-of-the-art performance in both cuboidal and non-cuboidal/generic layout datasets.  
	\item We propose a non-cuboidal layout parametric representation able to handle increasingly complex layouts. For this, we explicitly learn to predict its complexity parameters.
	\item We present a new way to generate layout candidates based on multiple kinds of feature maps, predicted layout complexity, and visual cue alignment.
	\item We are, to the best of our knowledge, the first to propose using a deep neural network to rank candidate layouts. For this, we introduce a new framework to train a CNN using max-margin structure loss.
\end{compactenum}

\section{RoomStructNet for layout estimation}
 \label{sec:RoomStructNet}
Our proposed system RoomStructNet takes an indoor scene image as an input, and estimates its non-cuboidal layout. We are able to handle complex non-cuboidal rooms thanks to our generic layout representation, detailed in Section~\ref{sec:representation}.

We show a diagram of our full pipeline in Figure~\ref{fig: whole pipeline}.
In the first stage, RoomStructNet uses a network to detect robust layout features as confidence maps that encode layout boundaries, corners, and surface segmentations. This is shown by black arrows in Figure~\ref{fig: whole pipeline}, and discussed in Section~\ref{sec:features}. 
In the next stage, RoomStructNet uses the computed robust features to generate candidate layouts, shown by gray arrows in Figure.~\ref{fig: whole pipeline}. We present our layout candidate generation methodology in Section~\ref{sec:Proposing layout}. 
In the last stage, RoomStructNet embeds the candidate layouts jointly with their robust features, and scores them to select the final estimated layout. This process is shown by blue arrows in Figure~\ref{fig: whole pipeline}, and described in Section~\ref{sec:ranking}.
 



%% file: sec_representation.tex
\subsection{Generic layout representation} \label{sec:representation}

Following the seminal work of Hedau et al.~\cite{Hedau}, most approaches attempt to estimate boxy layouts that segment an image into wall ($w$), ceiling ($c$) and floor ($f$) polygons. This is typically done by either sampling Manhattan rooms using orthogonal vanishing points, or relying upon 11 types of room layout configurations specified in the LSUN dataset~\cite{yu2015lsun}. Unlike previous works, here we propose a representation that makes no grossly simplifying assumptions about the captured room, such as perpendicularity of walls, the number of existing walls, or limited predefined room configurations. We only assume that all walls are vertical to the floor.

Given an image $I \in \mathbb{R}^{m\times n \times 3}$, we represent a layout $L$ containing an arbitrary number of walls $n$ as an ordered collection of wall-wall ($ww$) boundaries, which are defined by line segments $L=\{l_0^{ww}, l_1^{ww},\ldots,l_n^{ww}\}$. These line segments should all ideally intersect at the vertical vanishing point lying outside the image. Each segment $l_i$ joins a wall-wall-floor ($wwf$) corner $(x_i^f,y_i^f)$ with a wall-wall-ceiling ($wwc$) corner $(x_i^c,y_i^c)$. These corners can lie either inside the image if visible, or at image boundaries if not visible in the image. 
To account for $ww$ boundaries for walls not fully visible in the image, we allow $l_0$ to lie on the left image boundary and $l_n$ to lie on the right image boundary. Given the $ww$ line segments, the $i$-th $wf$ segment $l_i^{wf}$ is computed by joining $(x_i^f,y_i^f)$ and $(x_{i+1}^f,y_{i+1}^f)$. 
The floor polygon is computed by joining $wwf$ corners and the image boundaries. The $wc$ boundaries and ceiling polygon can be constructed in a similar fashion. As a dual representation for layout $L$, we create a 3-channel layout boundary image $I^L$ with each channel encoding one of $ww$, $wf$ or $wc$ boundaries.

%% file: sec_features.tex
\subsection{Robust layout features extraction} \label{sec:features}

Robust features are fundamental for correctly localizing candidate room layouts. As suggested by the success of recent works~\cite{PIONet, Mallya,LayoutNet,RoomNet}, we advocate that these features are better learned from data.

 Consequently, we propose using three kinds of such features together: boundary, corner and segmentation maps. We believe these complement each other in terms of information content and robustness. For instance, visible wall-wall and wall-ceiling boundaries are usually best captured by boundary confidence maps. While wall-wall-ceiling corners are best captured by corner confidence maps. Finally, occluded wall-floor boundaries and wall-wall-floor corners are best captured by analyzing the masks of occluding furniture items in segmentation maps. 
 
 Beyond estimating the aforementioned visual layout features, we propose using an additional network branch to predict the complexity of the scene depicted in the input image. We define room layout complexity as the number of visible walls in the scene. Since we perform semantic segmentation rather than instance segmentation, our segmentation maps do not separate different walls. Therefore, the predicted number of visible of walls nicely complements the segmentation information. 

Our proposed RoomStructNet feature extraction module takes an image as an input, and predicts three per-pixel maps and one value representing layout complexity level (see Figure~\ref{fig: whole pipeline}). The corner map, $M_c$, is a 2-channel image with pixel values expressing the confidence of detecting  $wwf$ and $wwc$ corners, respectively, in each channel. The boundary map, $M_b$, is a 4-channel image where pixel values in each channel denote, respectively, the confidence of detecting $ww$, $wf$, and $wc$ boundaries as well as non-boundary locations. The segmentation map, $M_s$, is a 3-channel image with each pixel labeled as wall $w$, floor $f$, or ceiling $c$. These labels are the result of a layout segmentation task, which ignores the occluding objects like furniture and labeling as if the room was empty.
 The module also outputs the predicted layout complexity level as a scalar $\mathcal{C}$, with $\mathcal{C}\geq 0$.  
 
 To extract these features we use a Dilated Residual Network (DRN)~\cite{yu2017dilated} with a configuration of 105 layers as our backbone network. To the network branches for corner, boundary, and segmentation prediction tasks, we add two convolutional layers with batchnorm and softmax in our network design. To the layout complexity level prediction branch, we add a fully connected layer at the end, which outputs the predicted number of visible walls. 

We train our proposed feature extraction network in two stages. First, we train jointly the DRN-105 backbone and the subsequent network branches of all four tasks until convergence. Second, we fix parameters of the DRN-105 backbone, and follow by training each branch separately with a smaller learning rate. We employ $L_2$ regression loss for the layout complexity prediction task and binary cross entropy loss for the other three tasks. The total loss of the network is the sum of the losses of all four tasks. 

%% file: sec_proposals.tex
\subsection{Generic layout candidate generation} \label{sec:Proposing layout}
We pose the problem of layout estimation as an object detection task, where the goal is to detect the layout that best suits the input image.
Deep learning based object detection methods can be broadly categorized into two groups. First, we have methods that pose object detection as a regression problem and adopt a unified framework to achieve the goal of either classifying or localizing objects~\cite{redmon2016you, liu2016ssd, najibi2016g}. And then, we have methods that follow the traditional object detection pipeline. That is, first target proposals are produced, and then each proposal is either classified or rank each into different object categories~\cite{girshick2014rich,he2015spatial,girshick2015fast,he2017mask}. Our proposed approach belongs to the second group: we first generate candidate layouts, and then we rank them to obtain the final layout estimate. 


The goal of our layout candidate generation is to propose wide range of plausible generic layouts that can potentially fit the image with high accuracy. And we do this without the assumption of cuboidal room layout.
To achieve this goal, we adopt a two stage process. In the first stage, we initialize all possible layouts for the given predicted layout features and complexity level. In the second stage, we use different visual cues to fine-tune the initial layout candidates and increase the accuracy of their localization. 

\subsubsection{Candidate layout initialization} \label{sec:Initializing layouts}
In contrast to many previous works~\cite{hirzer2020smart,Hedau,DasGupta}, we do not assume that the layout is cuboidal, and hence we cannot initialize the layouts using a limited number of predefined layout templates. We instead rely upon the signals provided by layout features and complexity level predicted by our network described in Sec.~\ref{sec:features} to initialize layouts.

During the training of our feature extraction network, we noticed that our prediction of boundary map $M_b$ and layout complexity level $\mathcal{C}$ is very precise for most cases -- the prediction overestimates the number of walls in 16\% of cases but never underestimates the number. So our initialization process uses predicted boundary as a starting point and produces room layout candidates with at most $\mathcal{C}$ walls. We consider all possible combinations of presence of ceiling, walls and floors in an image during the initialization, which includes $\{c,w,f\}, \{c,w\}, \{w,f\}$ and $\{w\}$.

To begin with, our initialization process detects $ww$ boundary candidates by using the boundary map $M_b$. Since $ww$ boundaries are mostly visible and crisp in input images, line detection on $M_b$ is straight forward. To detect these boundaries, we use image thresholding followed by morphological operation and line fitting on $M_b$.

\begin{figure}[th]
\centerline{
\begin{tabular}{c}
  \resizebox{0.47\textwidth}{!}{\rotatebox{0}{
  \includegraphics{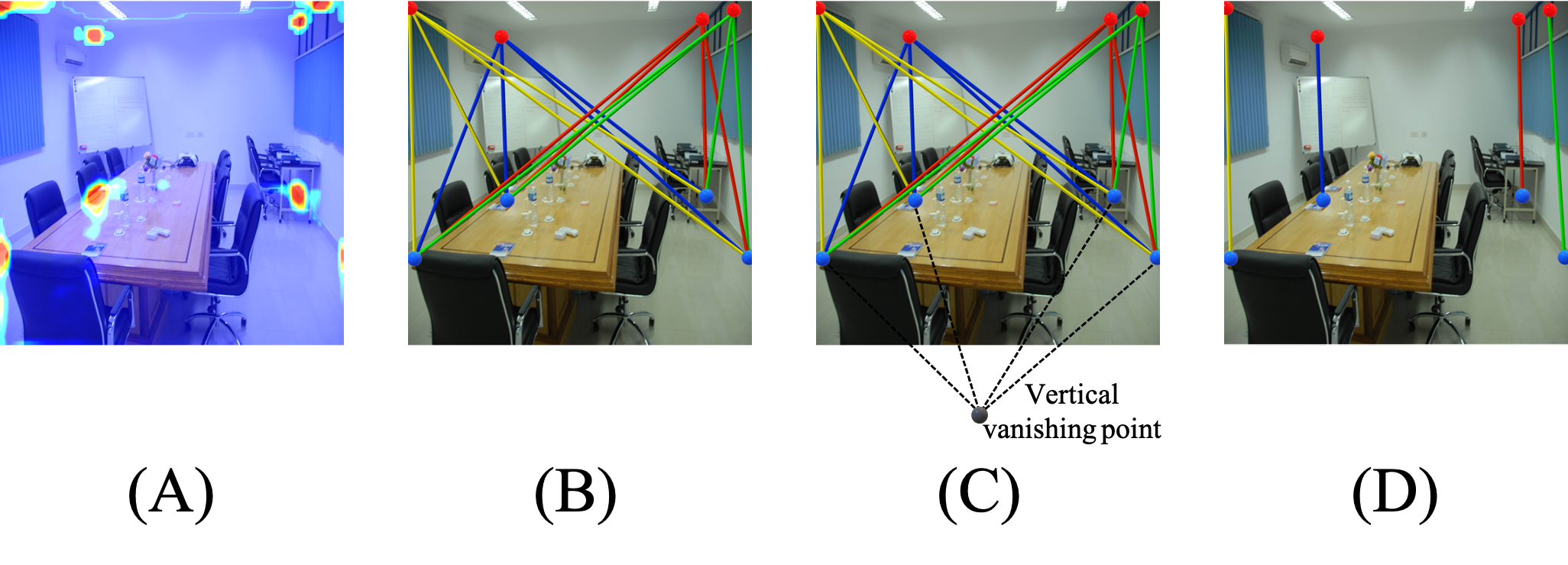}}}
  \\
\end{tabular}}
\caption{An illustration of the wall-wall boundaries proposing process. Given a predicted corner map, as shown in (A), we extract all local maximum as corner candidates and construct bipartite connections between the ceiling corners and the floor corners, as shown in (B). We then compare each ceiling-floor connection with the vertical vanishing point, shown in (C), and select the final wall-wall boundaries that follow the vanish point constraint, shown in (D).  }
\label{fig: corner wwlines}
\end{figure}

To guarantee high recall for later stages, we also extract additional boundaries from the corner map $M_c$. For this, we first compute the vanishing points of the image corresponding to three principal orthogonal directions using approach introduced in~\cite{lu20172}. Since we are not assuming the layout to be cuboidal, the vanishing points corresponding to horizontal room planes are not useful for us. We however assume the room walls to be vertical, hence we trust and use the vertical vanishing point in our work. Each pixel of our corner map $M_c$ describes the confidence of detecting one of ${wwf, wwc}$ corner types. We thus detect $wwf$ corners and $wwc$ corners by finding local maxima in $M_c$. We then create bipartite connections between every $wwf$ and $wwc$ corner pair. To select valid $ww$ boundaries, we examine alignment between each $ww$ boundary and the vertical vanishing point. When vanishing point is not successfully computed, the $ww$ boundaries are compared with vertical direction. Figure~\ref{fig: corner wwlines} shows this process of extracting lines from $M_c$.

\begin{figure}[htb]
\centerline{
  \begin{tabular}{cc}
    \resizebox{0.12\textwidth}{!}{\rotatebox{0}{
    \includegraphics{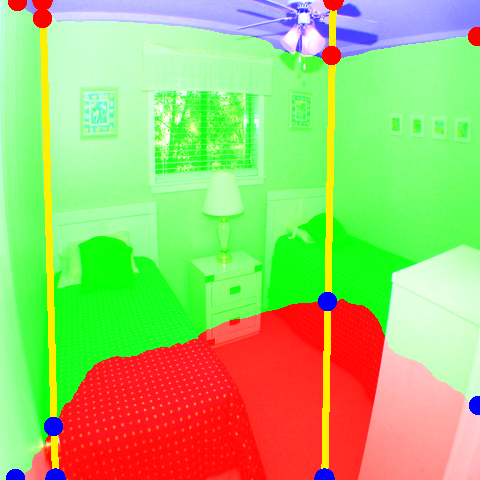}}}
    &
    \resizebox{0.12\textwidth}{!}{\rotatebox{0}{
    \includegraphics{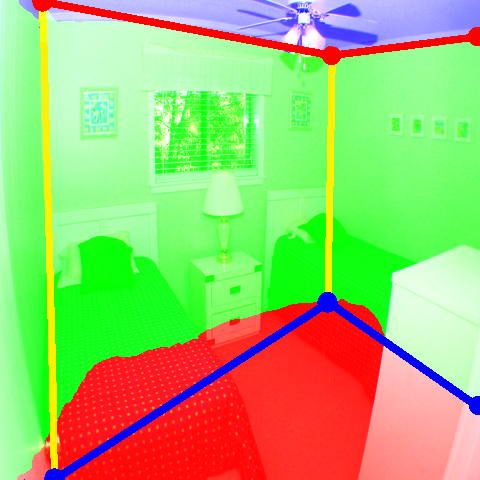}}}
    \\
    (A) & (B)
\end{tabular}}
\caption{An examle of layout initialization using $ww$ boundaries and $M_s$ segmentation mask. (A) First, we compute the intersections between $ww$ and $M_s$ mask. If any intersection is too close to an image boundary, we will set it to the intersection of $ww$ and the same boundary. (B) We initialize a layout by connecting ceiling and floor intersections. }
\label{fig: Initialize layouts}
\end{figure}

We then generate layout candidates using the detected $ww$ boundaries. Suppose we detect $N$ $ww$ boundary candidates from the maps $M_b$ and $M_c$ using the approach decribed above. To propose a layout containing $\mathcal{C}$ walls, we first select $\mathcal{C}-1$ of the $N$ $ww$ boundary candidates. We then find the top and bottom point for each $ww$ boundary line segment, corresponding to the $wwc$ and $wwf$ corner, using the floor and ceiling regions of the surface segmentation mask $M_s$.
We illustrate this process in Figure~\ref{fig: Initialize layouts}.
First we find the intersection between each $ww$ line and region boundaries in $M_s$. If any intersection is too close to an image edge, we will instead replace it with the junction between $ww$ and the image boundary. Finally, we connect the ceiling and floor points in order to form ceiling and floor regions respectively. At the end of the initialization process we obtain many layout candidates.

\subsubsection{Visual cue alignment for layout candidate refinement} \label{sec:optimizing layouts}
So far we have obtained a collection of initialized layouts $\textbf{L} = \{L_i\}_{i=0}^n$ of the given input image. Many previous works conduct an optimization process that iteratively move layout connection joints to achieve better score and select the layout with best score as the final output. 

Although an optimization process can improve layout accuracy, it also comes with drawbacks. First, it is computationally expensive and sometime requires a heuristic stopping criteria. Secondly, many works use the predicted feature map as the optimization space, which often leads to local minimas due to lack of actual visual signals.

\begin{figure}[htb]
\centerline{
  \begin{tabular}{ccc}
    \resizebox{0.14\textwidth}{!}{\rotatebox{0}{
    \includegraphics{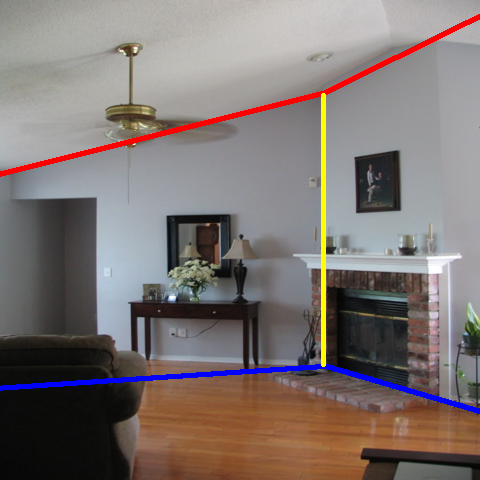}}}
    &
    \resizebox{0.14\textwidth}{!}{\rotatebox{0}{
    \includegraphics{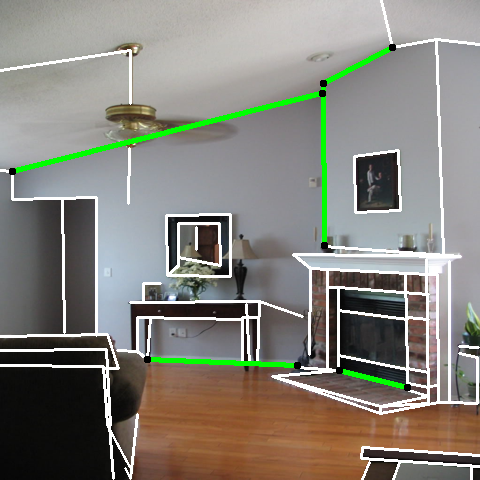}}}
    &
    \resizebox{0.14\textwidth}{!}{\rotatebox{0}{
    \includegraphics{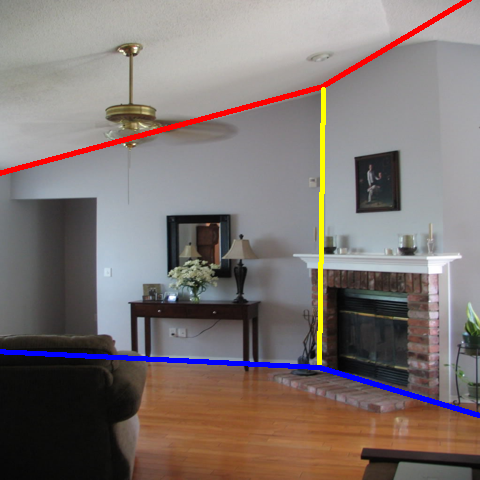}}}
    \\
    (A) & (B) & (C)
\end{tabular}}
\caption{An example of the visual cues alignment process. Given an initial layout propose in (A), we generate an layout aligned to the  visual cues in (C). We show all computed visual cues in (B) and highlight the line segments used for alignment in green. }
\label{fig: layout alignment}
\end{figure}

Instead of executing an iterative optimization process, we refine the candidates by aligning layout lines to visible cues in the scene, as illustrated in Figure~\ref{fig: layout alignment}. We first extract all visible line segments from the input image, and then align the initial layout boundaries to nearby line segments. We log all intermediate layouts and treat them as layout candidates as well.

Our refinement procedure should in theory produce similar or better results compared to an optimzation-based approach.  For scenes without occlusion, the layout initialization described in the previous section can already generate accurate layouts without additional refinement. For scenes with completely occluded layout boundaries and corners, an optimization process lacks additional signal compared to the feature extraction network. For scenes with partial occlusions, since some parts of layout lines and corners are still visible and detectable, our alignment-based approach has direct signals to improve the layout candidates. Our experiment results also validate our hypothesis here.

%% file: sec_ranking.tex
\subsection{Candidate layout scoring using CNNs} \label{sec:ranking}
Given an image $\mathrm{I}$ and a layout $L_{i}$ proposed by the methodology described in Sec.~\ref{sec:Proposing layout}, we want to determine whether it is a correct layout for the image. One approach is to pose this task as classifying the layout as correct or not, but proposing the ground truth layout is difficult. Instead, like most prior works, we formulate this task as learning a scoring function $f_s(L_{i},\mathrm{I})$ that will assign a higher score to a layout more similar to the ground truth.

We illustrate the layout scoring process with the blue arrows in the network architecture in Figure.~\ref{fig: whole pipeline}.
To jointly embed the image and a layout, we take concatenated network outputs from one layer before the layout features (described in Section.~\ref{sec:features}) and append them with the layout boundary image $\mathrm{I}^{L_{i}}$ produced from the candidate layout. We then pass it through a network with two Atrous residual blocks~\cite{chen2018encoder} to generate the final layout score. To reinforce the information provided by the candidate layout, we again append $\mathrm{I}^{L_{i}}$ to the network output between the two Atrous residual blocks. Given a set of proposed layouts $\textbf{L} = \{L_{i}\}_{i=1}^{n}$, our system selects the layout with the maximum score as the output:
\begin{equation*}
L^*=\argmax_{L_{i} \in \textbf{L}}f_s(L_{i},\mathrm{I}).
\end{equation*}
 
\textbf{Training using max-margin structure cost.} Room layout is more than just a classification label; it is a complex construct containing structural information of the scene, such as location and orientation of layout lines and their intersections. In this structured space, layouts which are closer to the ground truth should be penalized less than those further away. Thus we use the max-margin structure cost inspired by~\cite{zhang2015improving,li2015maximum} to train our network, which forces the score of the ground truth layout $\hat{L}$ of an image $\mathrm{I}$ to be larger than that of another layout $L_{i}$ by a margin $\Delta(L_{i},\hat{L})$, a non-negative function computing the deviation between two layouts. In this setup, we aim to minimize the structure cost
\begin{equation*}
	E(L_i)=\max \left(0, f_s(L_{i},\mathrm{I}) + \Delta(L_{i},\hat{L}) - f_s(\hat{L},\mathrm{I}) \right).
\end{equation*}

For the cost function, we minimize the cost for all available layouts, as described in Eq.~\ref{eq:cost2}. In the original Support Vector formulation in~\cite{tsochantaridis2004support}, the authors minimize the cost while maintaining margin constraints by using only the layout that violates the margin constraint the most, as described in Eq.~\ref{eq:cost1}. In their setup, this drastically reduces the number of constraints and hence the linear system of equations to be solved in each iteration. However, for our neural network formulation this adds complexity of search to otherwise differential formulation. We thus chose to adopt Eq.~\ref{eq:cost2}, which leads to faster convergence and allows more training samples and constraints to be utilized.

\begin{eqnarray}
	C_1 &=& \max\nolimits_{i} E(L_i)   \label{eq:cost1}\\
    C_2 &=& \sum\nolimits_{i}E(L_i)\label{eq:cost2}
\end{eqnarray}

\textbf{Computing the margin $\Delta(,)$.} Our margin function computes the deviation of a layout $L_i$ from the ground truth layout $\hat{L}$ as a combination of a line-based deviation and area-based deviation, as described in Eq.~\ref{eq:margin}. 
The line based deviation computes the distance between the respective layout boundaries of the two layouts, by summing the pixel distance and angular distance, as described in Eq.~\ref{eq:margin_line}.
To compute the area based deviation, we first convert a layout to the floor/wall/ceiling area masks, denote as $\{M^f,M^w,M^c\}$ respectively, by constructing the wall, ceiling and floor polygons as described in Sec.~\ref{sec:representation}.
We then compute $D_{area}$ as defined in Eq.~\ref{eq:margin_area}, where the first term computes the area shrunk in prediction compared with the ground truth, and the second term computes Intersection-over-Union (IOU).
In practice, we only consider the floor mask in the computation because floor affordance is more important in most applications.
\begin{eqnarray}
	&\Delta(L_i,\hat{L})=D_{line}(L_i,\hat{L}) + D_{area}(L_i,\hat{L}) \label{eq:margin}\\
	&D_{line}(L_i,\hat{L}) = \|L_i,\hat{L}\|^2 + \cos(L_i,\hat{L}) \label{eq:margin_line}\\
&D_{area}(L_i,\hat{L})  = \sum\limits_{\alpha \in \{f, w, c\}} 2-\frac{|M_i^{\alpha} \cap \hat{M}^{\alpha}|}{|\hat{M}^{\alpha}|} -\frac{|M_i^{\alpha} \cap \hat{M}^{\alpha}|}{|M_i^{\alpha} \cup \hat{M}^{\alpha}|} \label{eq:margin_area}	
\end{eqnarray}

%% file: sec_results.tex
\section{Experiments} \label{sec:results}
In this section, we show our experiment setup and discuss results of the proposed approach.

\subsection{Datasets} \label{sec:datasets}
We evaluated the proposed approach on two challenging room layout benchmarks: Hedau dataset~\cite{Hedau} and LSUN scene understanding dataset~\cite{yu2015lsun}. Hedau dataset consists of 209 training images and 105 testing images. LSUN is a relatively larger dataset, consisting of 4000 training images, 394 validation images, and 1000 testing images. Both benchmarks assume cuboidal layouts in images. Since Hedau dataset contains too few training images, we did not train the network separately on this dataset. We instead followed the previous works~\cite{Zhang,IM2CAD,PIONet} and performed testing on both LSUN and Hedau datasets using the model trained only on LSUN dataset.

\begin{figure}[th]
\centerline{
\begin{tabular}{c}
  \resizebox{0.36\textwidth}{!}{\rotatebox{0}{
  \includegraphics{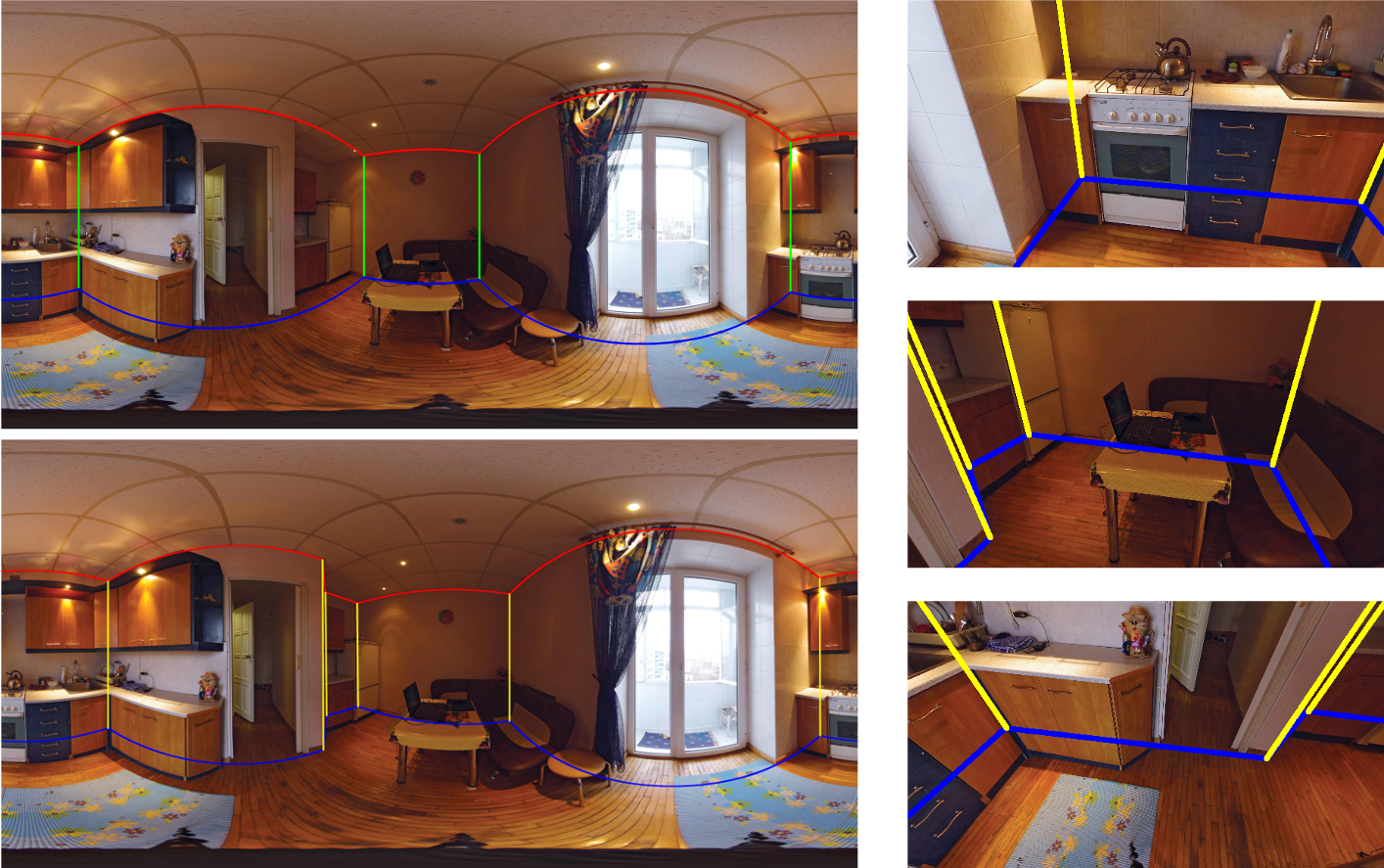}}}
  \\
\end{tabular}}
\caption{An example of a relabeled panorama and perspective views sampled from it. Upper left: the original labeling for the panorama. Bottom left: our retouched labeling. Right: perspective views sampled from the panorama.}
\label{fig: panoLayout}
\end{figure}

To show that the proposed approach can predict generic non-cuboidal layouts, we also evaluated the approach on a new room layout dataset that we created using PanoContext dataset~\cite{zhang2014panocontext}. The original PanoContext dataset consists of 700 full-view panoramas with 418 bedrooms and 282 living rooms. The original annotations make the Manhattan world assumption ~\cite{zou2021manhattan} and  contain only cuboidal layouts. To generate the new dataset, we selected 526 panoramas from the original dataset and relabeled the dataset without the Manhattan and cuboidal layout assumption. Then given a relabeled panorama, which covers 360 degrees in longitude and 180 degrees in latitude, we randomly sampled 6 perspective views every 24 degrees in longitude with FOV uniformly distributed between 70 and 120 degrees. Using this process, we generated $47,340$ perspective views with layout labels, and we splitted panarama views to training and testing sets and generated $40,000$ training and $7,340$ testing perspective views.  Examples of relabeled panorama and sampled perspective views are shown in Figure~\ref{fig: panoLayout}.

\begin{table}[ht!]
 \begin{center}
  \begin{tabular}{lcc}
    \toprule
    Method & ${\mathbf{e_{pixel}}}$  &  ${\mathbf{e_{corner}}}$ \\
    \midrule
 	Hedau $\etal$. (2009)~\cite{Hedau} &	24.23  &	 15.48 \\
 	Mallya $\etal$. (2015)~\cite{Mallya} & 16.71 & 11.02 \\
 	Dasgupta $\etal$. (2016)~\cite{DasGupta} & 10.63 & 8.20 \\
 	Ren $\etal$. (2016)~\cite{Ren} & 9.31 & 7.95 \\
 	Hsian $\etal$. (2019)~\cite{flat2Layout} & 6.68 & 4.92 \\
 	Hirzer $\etal$. (2020)~\cite{hirzer2020smart} & 7.79 & 5.84 \\
 	 \midrule
 	Ours w/o visual cue alignment & 7.32 & 5.03 \\
    RoomStructNet  &	\textbf{6.21}  & \textbf{4.65}  \\
    \bottomrule
  \end{tabular}
  \caption{Quatitative results on LSUN. The proposed approach achieved the best results compared to both existing approaches and the proposed approach without using visual cue alignment.}
   \label{tab:LSUN results}
   \end{center}
\end{table}

\subsection{Evaluation results} \label{sec:evaluation}
Similar to most other works, we evaluated the performance of our approach using the following two metrics:

\begin{compactitem}
  \item ${\mathbf{e_{pixel}}}$: Ratio of pixelwise error between predicted ${w, c, f}$ areas and ground truth areas, normalized by the image size.
  \item ${\mathbf{e_{corner}}}$: Ratio of euclidean distance between positions of predicted layout joints and positions of ground truth joints, normalized by the image diagonal.
\end{compactitem}

We present the results of the proposed approach and other approaches on LSUN dataset in Table~\ref{tab:LSUN results} and Hedau dataset in Table~\ref{tab:Hedau results}. For LSUN dataset, the ground truth testing set is no longer available on the official evaluation kit. We thus evaluated only on the validation set (also the case with \cite{RoomNet, Ren,  LayoutNet}). To verify and highlight the performance gain achieved by visual cue alignment, we conducted an ablation study by skipping the visual cue alignment in the proposed workflow (results also included in Table~\ref{tab:LSUN results} and Table~\ref{tab:Hedau results}). Note that we excluded \cite{PIONet} in our comparison, as it used training data not provided in the benchmark and is thus not directly comparable to the other works.

\begin{table}[ht!]
 \begin{center}
  \begin{tabular}{lc}
    \toprule
    Method & ${\mathbf{e_{corner}}}$ \\
    \midrule
 	Hedau $\etal$. (2009)~\cite{Hedau} &	21.2  \\
 	Mallya $\etal$. (2015)~\cite{Mallya} & 12.83 \\
 	Dasgupta $\etal$. (2016)~\cite{DasGupta} & 9.73 \\
 	Ren $\etal$. (2016)~\cite{Ren} & 8.67  \\
 	Hsian $\etal$. (2019)~\cite{flat2Layout} & 5.01 \\
 	Hirzer $\etal$. (2020)~\cite{hirzer2020smart} & 7.44 \\
 	 \midrule
 	Ours w/o visual cue alignment & 6.55 \\
    RoomStructNet  &	\textbf{4.81} \\
    \bottomrule
  \end{tabular}
  \caption{Quantitative results on Hedau dataset. }
   \label{tab:Hedau results}
   \end{center}
\end{table}

\begin{figure}[th]
\centerline{
\begin{tabular}{ccccccc}
  \resizebox{0.08\textwidth}{!}{\rotatebox{0}{
  \includegraphics{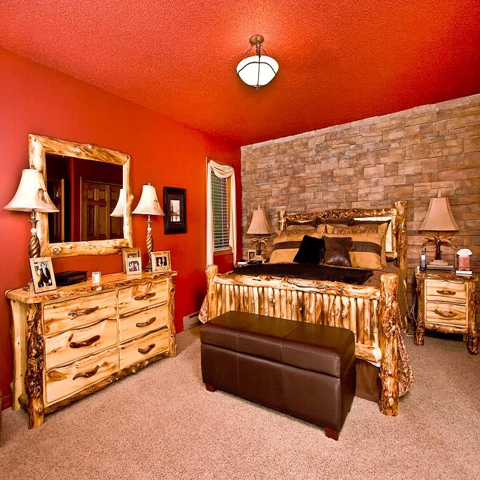}}}
  &
  \resizebox{0.08\textwidth}{!}{\rotatebox{0}{
  \includegraphics{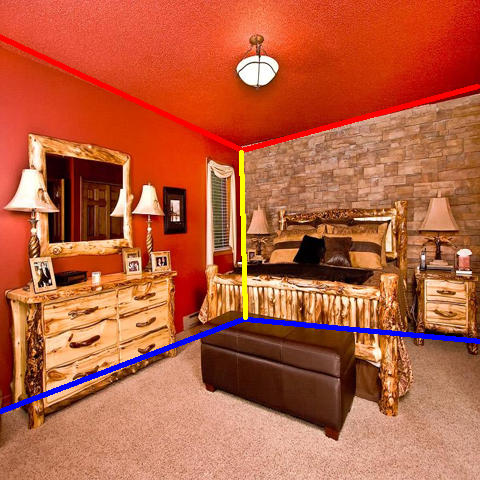}}}
  &
  \resizebox{0.08\textwidth}{!}{\rotatebox{0}{
  \includegraphics{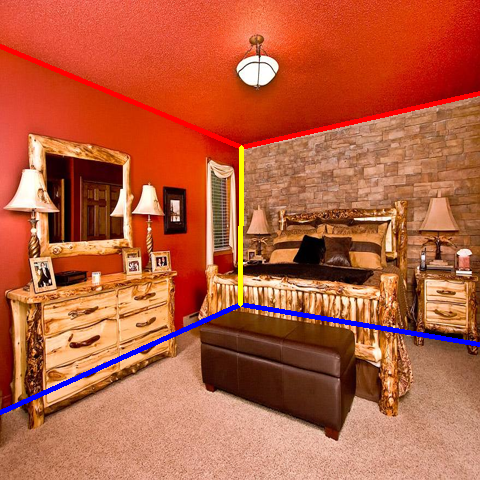}}}
  &
  \resizebox{0.08\textwidth}{!}{\rotatebox{0}{
  \includegraphics{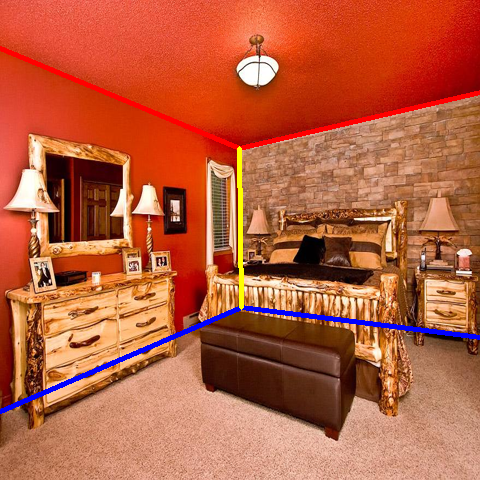}}}
  \\
  \resizebox{0.08\textwidth}{!}{\rotatebox{0}{
  \includegraphics{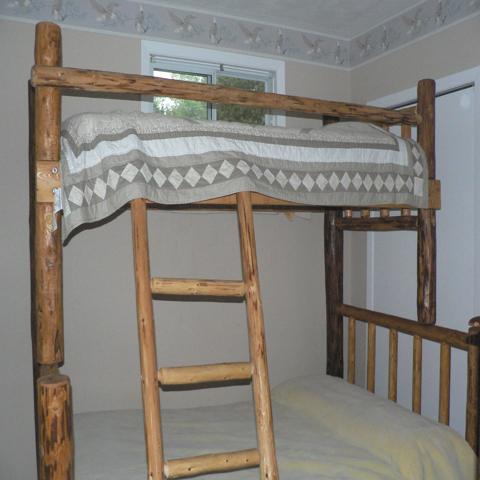}}}
  &
  \resizebox{0.08\textwidth}{!}{\rotatebox{0}{
  \includegraphics{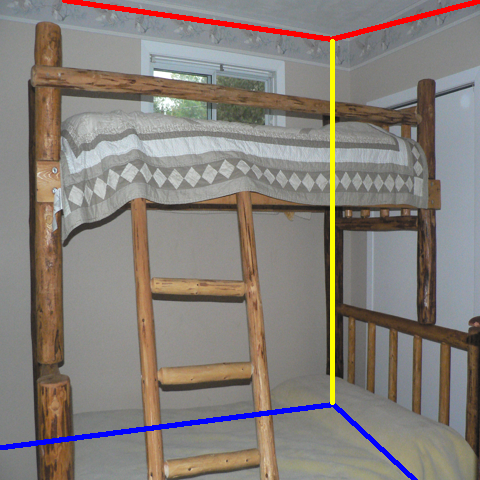}}}
  &
  \resizebox{0.08\textwidth}{!}{\rotatebox{0}{
  \includegraphics{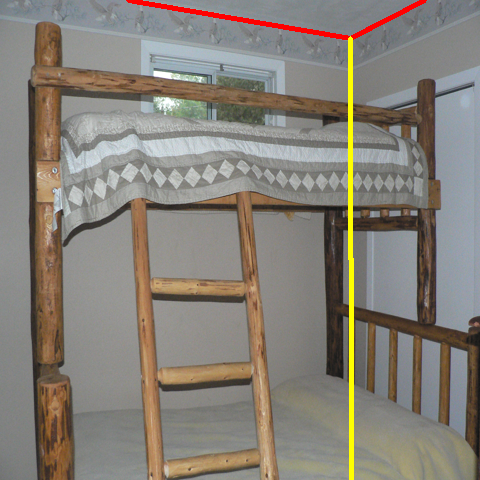}}}
  &
  \resizebox{0.08\textwidth}{!}{\rotatebox{0}{
  \includegraphics{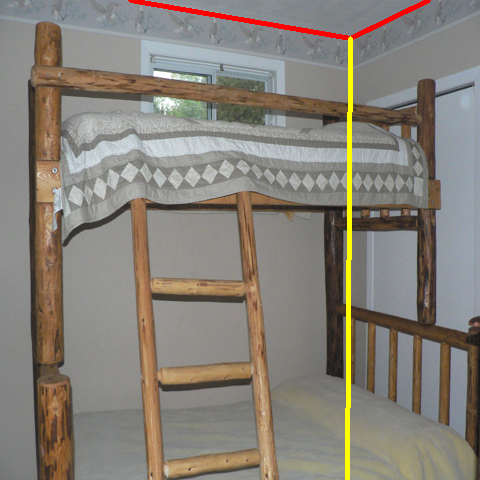}}}
  \\
  \resizebox{0.08\textwidth}{!}{\rotatebox{0}{
  \includegraphics{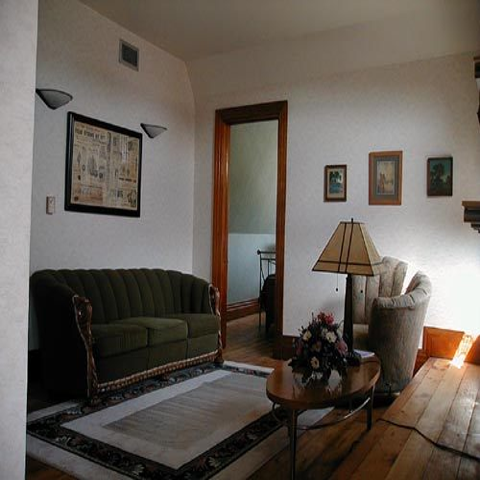}}}
  &
  \resizebox{0.08\textwidth}{!}{\rotatebox{0}{
  \includegraphics{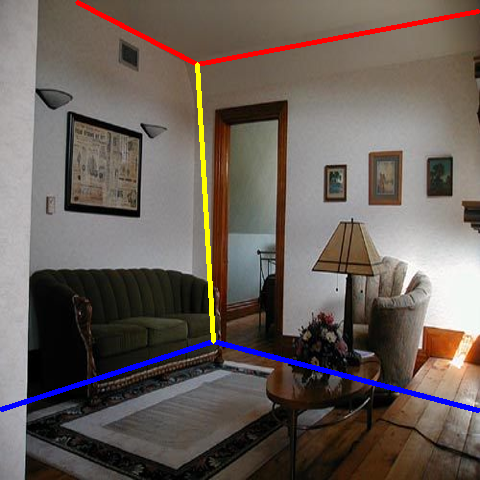}}}
  &
  \resizebox{0.08\textwidth}{!}{\rotatebox{0}{
  \includegraphics{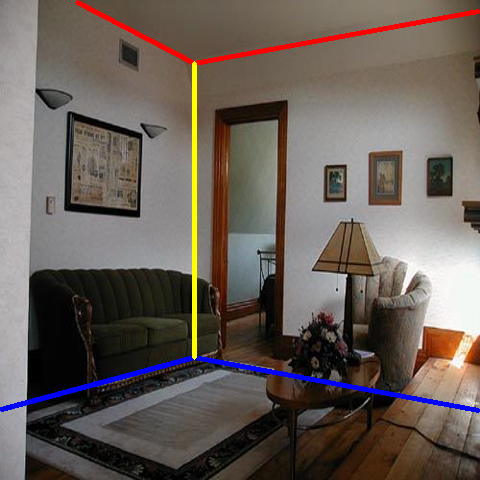}}}
  &
  \resizebox{0.08\textwidth}{!}{\rotatebox{0}{
  \includegraphics{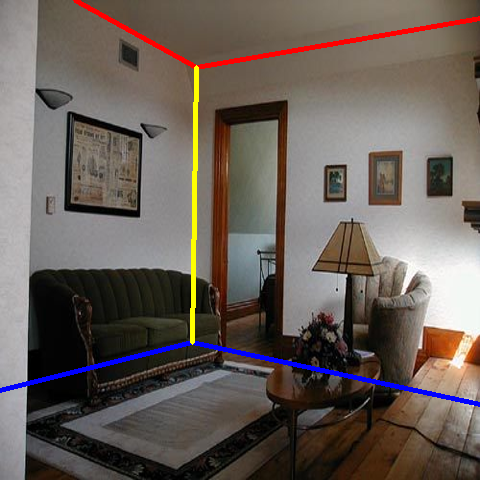}}}
  \\
  Input & w/o visual cue & Proposed & GT layout
\end{tabular}}
\caption{Qualitative results of the proposed approach on LSUN dataset.}
\label{fig: LSUN qualitative}
\end{figure}

We also show the qualitative results of the proposed approach on LSUN dataset in Figure~\ref{fig: LSUN qualitative}. It performs well in highly cluttered scenes.

\subsection{Results on generic room images}
\begin{figure}[th]
\centerline{
\begin{tabular}{c}
  \resizebox{0.5\textwidth}{!}{\rotatebox{0}{
  \includegraphics{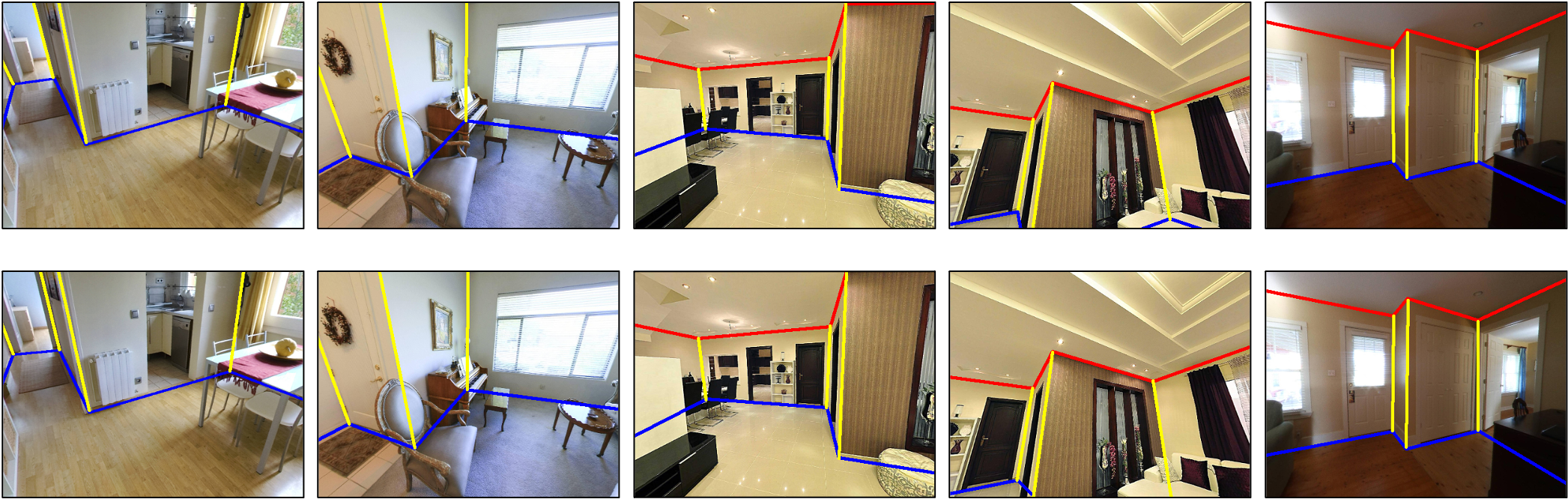}}}
  \\
\end{tabular}}
\caption{Qualitative results of the proposed approach on perspective views extarcted from the relabeled PanoContext dataset. It can be observed from the figure that the proposed approach can handle non-cuboidal layouts of the datasets.}
\label{fig: panoLayout qualitative}
\end{figure}

\begin{figure}[th]
\centerline{
\begin{tabular}{c}
  \resizebox{0.46\textwidth}{!}{\rotatebox{0}{
  \includegraphics{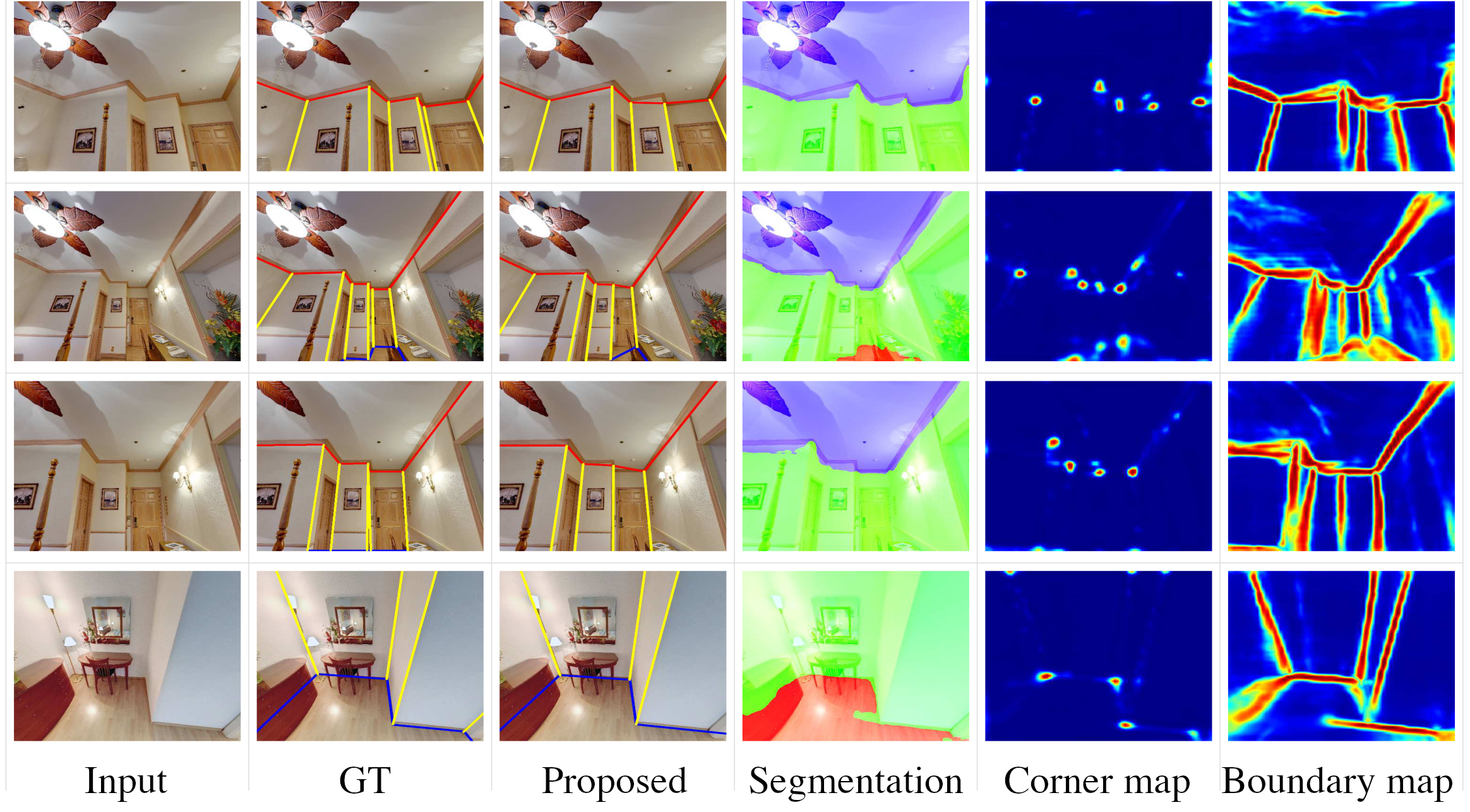}}}
  \\
\end{tabular}}
\caption{Examples of showing that the proposed layout feature extractor can detect layout features from rooms with generic layouts. }
\label{fig: layout features of generic rooms}
\end{figure}

One advantage of the proposed approach is that it can predict non-cuboidal room layouts. Many works addressed generic room layouts prediction in the past.  But they either assumed camera model exists in the data and used camera pose in their approaches~\cite{howard2018thinking,  RenderAndCompare} or conducted their prediction work on panoramic views~\cite{fernandez2020corners, zou2021manhattan,  pintore2020atlantanet}.  The only work we found has a similar problem setup as us is~\cite{flat2Layout}, however we could not find the code of their approach online. At this point, to demonstrate this, we evaluated the performance of the proposed approach on the perspective views derived from PanoContext dataset and present results here by using the same evaluation metrics.  We present some of qualitative results from our experiment in Figure~\ref{fig: panoLayout qualitative}.  We also present qualitative results of layout feature maps extracted from room with generic layouts in Figure~\ref{fig: layout features of generic rooms}. It can be observed from the feature maps that the proposed feature extractor is capable of crisply detecting and locating layout features of given rooms, which enables the capability of detecting the generic layouts in the proposed approach.  

\begin{figure}[th]
\centerline{
\begin{tabular}{ccccccc}
  \resizebox{0.5\textwidth}{!}{\rotatebox{0}{
  \includegraphics{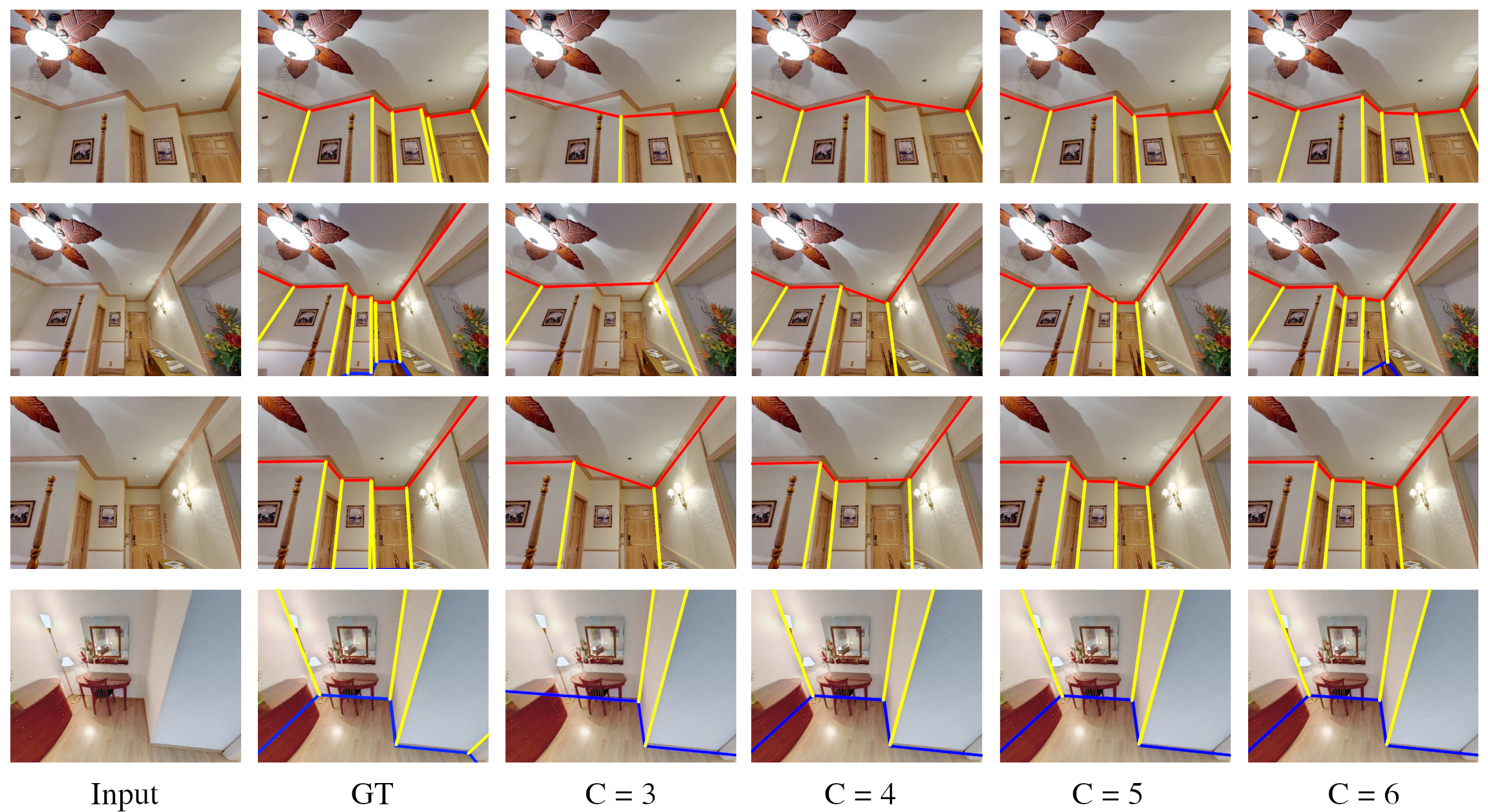}}}
  \\
\end{tabular}}
\caption{Qualitative results of setting different upper bounds for complexity factor $\mathcal{C}$.}
\label{fig: Complexity levels}
\end{figure}

We believe that the proposed complexity factor which controls the number of walls in predicted layouts plays an important role in generating final layouts.  To better understand the impact of the complexity factor to the predicted layouts. We conducted an evaluation in which we included versions of our proposed approach with $\mathcal{C}$ set to different numbers. Worth noting that by setting the complexity factor $\mathcal{C}$ to $3$, the proposed approach will only predict cuboidal layouts.  Table~\ref{tab:panoLayout results} summarizes the results described above.  We notice that the improvement becomes ignorable when $\mathcal{C}>6$.  This is because for most of room images in our relabeled panoContext dataset has the number of walls less than or equal to 6. The proposed approach achieves better performance thanks to its ability to predict generic layouts.  Some qualitative results of using different upper bounds for the complexity factor is shown in Figure~\ref{fig: Complexity levels}.  

\begin{table}[ht!]
 \begin{center}
  \begin{tabular}{lcc}
    \toprule
    Method & ${\mathbf{e_{pixel}}}$  &  ${\mathbf{e_{corner}}}$ \\
    \midrule
	RoomStructNet, $\mathcal{C}=3$ & 11.32 & 8.12 \\
 	RoomStructNet, $\mathcal{C}=4$ & 10.97 & 7.81 \\
 	RoomStructNet, $\mathcal{C}=5$ & 9.74 & 6.92 \\
 	RoomStructNet, $\mathcal{C}=6$ & 9.37 & 6.54 \\
    \bottomrule
  \end{tabular}
  \caption{Quatitative results on perspective views derived from PanoContext dataset.}
   \label{tab:panoLayout results}
   \end{center}
\end{table}

\subsection{Structure loss V.S. regression loss}
While the proposed approach achieves superior performance by training layout scoring network using max margin structure loss, an alternative way is to train the network as a regression function that fits the layout margin we defined in Eq.~\ref{eq:margin_line}. To compare, we trained the same network using the same LSUN layout candidate set extracted from the layout proposal process, with the loss function changed from structure loss to standard L2 loss. We then computed the average ${\mathbf{e_{pixel}}}$ (\%) for Top 1, 5, 10, 20 layouts ranked by the trained networks.
We examined multiple top layouts because a good layout scorer should still select a high quality candidate even when it does not select the best one.
We hypothesize that the network trained with the structure loss can better achieve this desired property as it learns the order relationship among input layouts.
The experiment results, summarized in Table~\ref{tab:L2 loss compare}, validates our hypothesis. It shows that the error rate for the network trained with the strcuture loss increases more slowly than the one with L2 loss.

\begin{table}[ht!]
 \begin{center}
  \begin{tabular}{lcccc}
    \toprule
    Loss type & Top 1  &  Top 5 & Top 10  & Top 20 \\
    \midrule 
    L2 loss 			   & 4.82 & 5.26 & 6.94 & 9.52 \\
	Structure loss & 4.65 & 4.82 & 6.71  & 9.52 \\
    \bottomrule 
  \end{tabular}
  \caption{A comparison between layout scoring networks trained using two different loss functions. We report the average ${\mathbf{e_{pixel}}}$ (\%) for the top-k layouts ranked by the trained network on LSUN dataset.}
   \label{tab:L2 loss compare}
   \end{center}
\end{table}

%% file: sec_conclusion.tex
\section{Conclusion} \label{sec:conclusion}
In this paper, we present a new approach to estimate the layout of a room from its single image. While most of the recent approaches for this task use robust features learnt from data, they resort to optimization when detecting the final layout. In addition to using learnt robust features, the proposed approach learns an additional ranking function to estimate the final layout instead of using optimization. To learn this ranking function, we propose a framework to train a deep neural network using max-margin structure cost. Also, while most approaches aim at detecting cuboidal layouts, our approach detects non-cuboidal layouts for which we explicitly estimates layout complexity parameters. We use these parameters to propose layout candidates in a novel way. Our approach shows state-of-the-art results on standard datasets with mostly cuboidal layouts and also performs well on a dataset containing rooms with non-cuboidal layouts.